# Blind Separation of Vibration Sources using Deep Learning and Deconvolution

Igor Makienko[a], Michael Grebshtein[a], Eli Gildish[a] *

[a]*RSL Electronics LTD, Migdal Ha'Emek, 23100, Israel*

**Abstract**

Vibrations of rotating machinery primarily originate from two sources, both of which are distorted by the machine's transfer function on their way to the sensor: the dominant gear-related vibrations and a low-energy signal linked to bearing faults. The proposed method facilitates the blind separation of vibration sources, eliminating the need for any information about the monitored equipment or external measurements. This method estimates both sources in two stages: initially, the gear signal is isolated using a dilated CNN, followed by the estimation of the bearing fault signal using the squared log envelope of the residual. The effect of the transfer function is removed from both sources using a novel whitening-based deconvolution method (WBD). Both simulation and experimental results demonstrate the method's ability to detect bearing failures early when no additional information is available. This study considers both local and distributed bearing faults, assuming that the vibrations are recorded under stable operating conditions.

## 1. Introduction

Traditionally, vibration signals are processed offline using robust computational resources, along with synchronized supplementary information such as system kinematics and operating conditions like rotational speed or load [1]. Transitioning to edge processing could potentially reduce communication traffic, lower offline computational expenses, and facilitate decision-making closer to the monitoring equipment. This study introduces a novel method that enables the separation and estimation of vibration sources at the edge, under the assumption that no information about system kinematics and operating conditions is available.

Analyzing vibration signals in geared machines, especially under fluctuating operational conditions, is inherently complex. As highlighted in [2], [3] and [4], the absence of comprehensive information about the machine, including its instantaneous rotational speed and load, presents a significant challenge. This study presumes that the machine's operational conditions are stable, with a focus on blind source separation. However, additional research is required to expand this study to accommodate variable conditions.

A typical vibration signal comprises two main sources, as detailed in reference [5]. The first source stems from subsystems such as gears phase-locked to the rotational speed, resulting in periodic (or quasi-periodic) deterministic signal, assuming stable operating conditions. The second source is stochastic, usually linked with bearings that experience random ball slippage, and measurement noise. The noise gathers data about the system's structural response as it propagates to the sensor [6]. Moreover, these two sources pass system's structural response en route to the sensor, a phenomenon termed here as the transfer function (TF) effect. This TF distorts the original information by modifying the signal amplitude at different frequencies [7].

* Corresponding author: (I. Makienko)
 *E-mail address:* ig.makienko@gmail.com; igor@rsl-electronics.com

Estimating vibration sources and eliminating the transfer function (TF) effect are important for monitoring gears and bearings. These steps can also serve as preprocessing in machine learning models of mechanical faults, as shown in recent studies [8],[9]. This is particularly relevant when historical data from identical machines is compiled for fault classification [10],[11] predicting the remaining useful life [12], or extracting valuable information from complex vibration signals [13].

The vibrations associated with bearing faults can represent both local and distributed faults, as illustrated in [5] and [14]. Their first two moments fluctuate periodically (or quasi-periodically) and are defined as second-order cyclostationary (or quasi-cyclostationary) (CS2) signals. This study addresses the problem of blind separating and estimating vibration sources without requiring additional information such as system kinematics or measured operating conditions.

The paper is structured as follows: Section 2 provides all the definitions and outlines the problem; Section 3 presents the related works in this field; Section 4 explains the method and provides a summary of the algorithm; Section 5 showcases evaluation results using simulations and discusses the selection of the algorithm's parameters; Section 6 concludes the paper and offers recommendations for future research.

## 2. Problem Statement

### 2.1. Definition1 - CS2

Define $c(t)$, a real-valued discrete time signal, where $t$ denotes discrete time samples. This signal is considered to be second-order cyclostationary in the wide sense with a period $T > 0$, if its mean and autocorrelation function are periodic (or quasi-periodic) in time with a period $T$, for all values of $t$ and $\tau$:

$$R_{cc}(t,\tau) \equiv \mathbb{E}\{c(t+\tau)c(t)\} = R_{cc}(t+T,\tau) \tag{1}$$

$$\mathbb{E}\{c(t+T)\} = \mathbb{E}\{c(t)\},$$

where $R_{cc}(t,\tau)$ is the autocorrelation of $c(t)$, $\mathbb{E}\{*\}$ denotes an expectation operator and $\tau$ signifies the time lag between distinct times. In a more general sense, the first two moments could be quasi-periodic, signifying a combination of periodic components with a least common period that could be infinite in time. The study addresses the general case, referring to the signals as CS2.

### 2.2. Definition2 – Bearing Faults

As detailed in [5] and further expanded in [40], the vibrations resulting from bearing defects can be characterized as CS2 signals, with the definition varying according to the fault type: local or distributed.

<u>Local faults</u> generate shocks each time the rolling components encounter the defect, with sliding introducing a random element to these shocks as follows:

$$c_l(t) = q(t) \sum_i e(t)\delta(t-T_i). \tag{2}$$

where $q(t)$ is a non-negative quasi-periodic function representing the amplitude modulation, $e(t)$ is a zero-mean white noise determining the randomness of the impact magnitude, $\delta(t)$ is the Dirac delta function, and $T_i$ is the random time of shock appearance whose difference $\Delta T = T_{i+1} - T_i$ is normally distributed.

This signal exhibits CS2 characteristics as follows:

$$\mathbb{E}\{c_l(t)\} = q(t) \sum_i \underbrace{\mathbb{E}\{e(t)\}}_{\equiv 0} \delta(t - T_i) = 0$$

$$R_{c_l c_l}(t, \tau) \equiv \mathbb{E}\{c_l(t) c_l(t - \tau)\}$$
$$= q(t)q(t - \tau) \sum_i \sum_j \mathbb{E}\{e(t - T_i)e(t - T_j - \tau)\} = \begin{cases} q^2(t)\delta_{ij}\sigma_e^2, & \tau = 0 \\ 0, & \tau \neq 0 \end{cases} \quad (3)$$

where $\delta_{ij}$ is the Kronecker delta, and $\sigma_e^2$ is the variance of the normally distributed $e(t)$.

For the underlined faults, the same $q(t)$ and white noise $e(t)$ are used for convenience. Here, $q(t)$ modulates the underlying white noise as follows:

$$c_d(t) = q(t)e(t). \quad (4)$$

The signal also demonstrates CS2 characteristics. The mean and auto-correlation of $c_d(t)$ are as follows:

$$\mathbb{E}\{c_d(t)\} = q(t) \underbrace{\mathbb{E}\{e(t)\}}_{\equiv 0} = 0$$

$$R_{c_d c_d}(t, \tau) \equiv \mathbb{E}\{c_d(t) c_d(t - \tau)\} = \begin{cases} q^2(t)\sigma_e^2, & \tau = 0 \\ 0, & \tau \neq 0 \end{cases} \quad (5)$$

at $\tau = 0$, it varies periodically over time due to the periodic nature of $q(t)$. Consequently, $c_d(t)$ possesses a zero mean and is uncorrelated.

As the actual fault type is not known a priori and both local and distributed faults exhibit similar characteristics, the distributed fault signal, denoted as $c(t) = q(t)e(t)$, will be used in the paper targeting the signal waveform estimation. Then, the method will be extended to accommodate both types of faults. The signal is CS2 with a zero mean and uncorrelated, where the waveform of $q(t)$ serves as the estimation target in this study.

## 2.3. Problem Definition

Consider a vibration signal $s(t)$, recorded from a geared system with bearings. This signal is composed of three sources that undergo the same linear and invertible TF before being measured by the sensor:

$$s(t) = \left( \underbrace{p(t)}_{gear} + \underbrace{c(t)}_{bearing} + \underbrace{w(t)}_{noise} \right) * \underbrace{h}_{TF\ filter} \quad (6)$$

where:
operator $*$ represents convolution.
$h$ is the TF filter of length $n$.
$p(t)$ is a deterministic quasi-periodic signal primarily related to gear vibrations.
$c(t) = q(t)e(t)$ is the zero mean and uncorrelated CS2 signal as defined in section 2.12.2. The signal only exists if a bearing fault appears.
$w(t)$ is a white measurement noise that is not correlated with $p(t)$ and $c(t)$.
According to this definition, the signal $s(t)$ consists of two components: a periodic signal $p(t)$ and a stochastic signal $c(t) + w(t)$. Both components pass through the TF $h$ on their way to the sensor.

The objective of this study is to estimate the waveforms of both $p(t)$ and $q(t)$, in the absence of additional information or measurements. Consequently, the problem of blind separation of vibration sources and estimation of their actual time waveforms is addressed.

## 3. Related works

The solution to the dual problem described in the previous section requires the utilization of two distinct categories of methods:
- Periodic-Stochastic Separation (PSS): This category enables the separation of gear-related periodic components and bearing-related stochastic components.
- Deconvolution techniques: These techniques are used to remove the TF effect, thereby facilitating the estimation of the actual signal waveforms.

An additional constraint is that both methods must operate independently of external measurements or additional system information.

*3.1. PSS methods*

A comprehensive review of the PSS methods is presented in [15]. The oldest technique, Time Synchronous Averaging (TSA), as described in [17] and [18], uses the machine's angular speed from a tachometer or encoder to perform angular resampling and transform the signal from the time domain to the angle domain. By averaging the phase-locked cycles, the periodic component is emphasized and can be separated from the stochastic part. However, this method requires machine angular speed for signal resampling, which is not available at sensor, and thus cannot be utilized in this study.

Another method, pioneered in [16], is a cepstrum-based separation of discrete components. The cepstrum is the inverse Fourier transform of a logarithmic spectrum, transforming the harmonics phase-locked to the shaft to known locations in quefrency (the frequency analog of cepstrum). These harmonics can be easily detected and isolated. This method works well with gear-related vibrations that need to be separated from other sources but requires information about system kinematics, which is unavailable in this study.

Linear prediction as in [19] and [35], doesn't require extra information and represents another PSS method. It assumes the autoregressive (AR) model of the signal and enables separation by utilizing a filter. The coefficients of this filter are calculated with Yule-Walker equations to minimize the error between the actual signal samples and those predicted using their close history. Unfortunately, the AR assumption is true only if the stochastic component of vibrations is white which is generally not the case since the background white noise passes through the TF on its way to the sensor, adding additional correlation between samples of the stochastic signal component.

The self-adaptive noise cancellation (SANC) method was proposed in [20]. The separating filter is built using the assumption that the stochastic and periodic signal parts have different correlation times. This assumption is realistic since the deterministic periodic part can be predicted using far remote historical samples, and the correlation of the stochastic part is limited by the length of the TF filtering both sources on their way to the sensor. This method fits the requirements since it doesn't require information about the signal or the monitored system. It uses only two main parameters: filter length and time delay. However, it has a drawback: the good separation capability in the frequency domain requires a very long filter length to ensure good frequency resolution. As a result, long stationary vibration recordings are required for the iterative algorithm to converge.

Recently, the use of a linear dilated CNN was proposed in [21] to solve this drawback. The method enables a significant increase in filter length while keeping the number of optimization parameters small. It was demonstrated in [21] that the dilated CNN requires significantly fewer optimization parameters and shorter signals compared to the SANC algorithm. This is important in practice, as machine operating conditions are constantly changing, making it challenging to capture stationary vibration signals over extended periods. Like SANC, this method doesn't require any additional information or extra measurements, aligning well with the requirements of this study.

In this study, the dilated CNN is employed for solving the PSS problem. Additionally, assuming the periodic component is eliminated, the second task involves conducting signal deconvolution to mitigate the TF effect, thereby enabling the estimation of the time waveform of the CS2 signal.

*3.2. Deconvolution Methods*

Numerical strategies for the deconvolution of vibration signals have been documented and recently consolidated in [32]. A subsequent study [22] extended on this, focusing on monitoring mechanical gear faults.

The most common technique is the Minimum Entropy Deconvolution (MED) method, first proposed in [23]. This filtering process simplifies the signal elements, which can be interpreted as decreasing the entropy value of the signal. Hence, the deconvolution is termed as minimum entropy deconvolution. The MED technique identifies the optimal filter coefficients that yield an inverse-filtered signal with maximum kurtosis. Methods based on MED eliminate the TF effect, assuming that the signal of interest should be impulsive.

A non-iterative adaptation of MED, termed MEDD [24], was introduced, featuring a novel statistical metric named D-Norm. D-Norm exhibits superior performance in scenarios with low Signal-to-Noise Ratio (SNR), while requiring fewer computational resources. MEDD provides a straightforward solution for deriving an optimal deconvolution filter. Subsequently, a more computationally efficient variant of MEDD, avoiding the matrix inversion, was proposed in [25]. The study [26] presented an alternative blind deconvolution approach where the optimal filter coefficients are determined through an objective function based on the Jarque–Bera statistic. Independently, the sensitivity of MED to intense single impulses and its filter length were investigated using a technique called $l_0$-norm embedded MED, introduced in [27]. This method aims to find a solution by employing an approximate hyperbolic tangent function for improved estimation of the $l_0$-norm. While many MED-based techniques address the TF effect by assuming the filtered signal should be impulsive, this approach may not be optimal for scenarios where the filtered signal represents a periodically modulated white noise, as addressed in this study.

Some methods leverage the assumption of signal sparsity for computing the coefficients of the deconvolution filter. A set of sparsity metrics, introduced in [28], utilizing the Box-Cox transformation [28], is employed within a sparsity-focused blind filtering framework utilizing Rayleigh quotient optimization, detailed in [29].

Another investigation aims to extract the cyclostationary signature inherent in the squared envelope spectrum by employing an optimal filter, whose estimation is accomplished through a gradient descent algorithm [30]. In a separate work [31], authors propose employing cyclostationarity indicators as an alternative to the maximum kurtosis criterion in deconvolution, labeling this approach as maximum second-order cyclostationarity blind deconvolution (CYCBD). This method iteratively addresses the new optimization problem through an eigenvalue decomposition algorithm.

Most techniques design the deconvolution filter relying on assumptions about specific properties of the target signal, such as periodic impulses or signal sparsity. This limits their applicability, especially when no additional information is available.

The deconvolution method in this study represents an improvement of the method [34] for deconvolution and waveform estimation of noisy CS2 signals validated through simulations under the assumption that the PSS problem was previously addressed. Here, we extend and improve the algorithm to work on any vibration signals and demonstrate its performance through simulations and real-world vibrations, including those resulting from bearing faults.

To summarize the review of PSS and the deconvolution methods, the blind separation and estimation of the vibration sources corrupted by the TF effect and without additional information or measurements is missing.

## 4. Methodology

*4.1. Solution Summary*

The method comprises four distinct stages outlined below:
1. Periodic-stochastic signal separation (PSS) utilizing dilated Convolutional Neural Networks (CNN).
2. Estimation of the deconvolution filter using the stochastic component.
3. Deconvolution of both the periodic and stochastic components, along with the estimation of $p(t)$.
4. Conducting statistical hypothesis testing to ascertain the presence of the CS2 signal and, if present, estimating

$q(t)$

### 4.2. Deterministic and Stochastic Components

Re-write the equation (6) as a combination of two components: periodic and stochastic, as depicted below:

$$s(t) = \underbrace{p(t) * h}_{deterministic} + \underbrace{(c(t) + w(t)) * h}_{stochastic} \quad (7)$$

$$\equiv d(t) + x(t)$$

where:
$d(t)$ and $x(t)$ represent the deterministic component and stochastic components, respectively, influenced by the TF. According to the World's theorem, it is guaranteed that the decomposition of signal $s(t)$ into its periodic $d(t)$ and stochastic $x(t)$ components is always feasible and exists.
The values of $d(t)$ can be accurately predicted based on distant past values. In contrast, the prediction accuracy of the stochastic component $x(t)$ decreases over time due to diminishing correlation. As a result, non-deterministic time-series should tend toward zero as the prediction time-lag increases, and the prediction error equals the signal itself. This separation technique was initially introduced in [20] and further extended in [37].
By employing a substantial time lag $\alpha$, ensures that for time lags $\tau \pm \alpha$, $x(t)$ becomes uncorrelated and $R_{xx}(t,\tau) = 0, \forall |\tau| > \alpha$ and $\forall t$. This implies that for time lags exceeding $\pm\alpha$, the model predicting $s(t)$ values should be identical to the model for $d(t)$. In this scenario, the deterministic signal $d(t)$ can be estimated as a linear combination of the past $m$ samples of $s(t)$:

$$\hat{d}(t) = \hat{s}(t) = \sum_{i=1}^{m} a_i s(t-i-\alpha), \quad (8)$$

$$\hat{x}(t) = s(t) - \hat{s}(t).$$

where coefficients $a_i$ denote the coefficients of the linear filter and $\alpha$ is the time-lag.

### 4.3. Periodic-Stochastic Separation (PSS)

The study addresses the blind PSS problem, aiming to distinguish between $d(t)$ and $x(t)$ without prior knowledge of the expected filter length. When $d(t)$ consists of multiple sinusoids with closely spaced frequencies, a long filter length $m$ (or receptive field) is required to achieve separation with sufficient frequency resolution. This results in a large number of optimization parameters in traditional methods such as SANC [21][20]. The use of dilated CNN, as recently suggested in [21], is advantageous in extending the receptive field while keeping the number of optimization parameters small. This benefit allows for optimization convergence with a shorter signal length. In this study, the architecture of the dilated CNN is assumed to be optimized, with the focus being on signal waveform estimation. Here is a brief overview of the proposed method.
To broaden the model's receptive field, its depth is increased, and weights are optimized through backpropagation. A linear dilated CNN is chosen, as the optimal predictor for the periodic components is expected to be linear, as demonstrated in equation (8). As outlined in [36], considering signal $s(t)$ and a dilated CNN with $L$ layers, the input at each layer is derived from the output of the preceding hidden layer and can be represented as follows:

$$u^{(l)}(t) = \beta^l *_\Delta u^{(l-1)}(t) = \sum_{k=1}^{m} \beta_k^l u^{(l-1)}(t - \Delta_l k). \quad (9)$$

Here, $u^{(l)}(t)$ denotes the output of layer $l$, the operator $*_\Delta$ indicates the dilated convolution, $\Delta_l$ and $\beta^l$ represent the dilation factor and weights (or kernel) of layer $l$ respectively and $m$ is the kernel size. Unlike regular convolution, in dilated convolution the filter is applied to every $\Delta_l$th element in the input vector. This allows the model to effectively learn connections between distant data points, facilitating the efficient capture of long-range dependencies within the input signals. We employ an architecture consisting of $L$ layers of dilated convolutions, with the dilation factor increasing by a factor of 2 in each subsequent layer: $\Delta_l \in [2^0, 2^1, .. 2^{L-1}]$. An example of a three-layer dilated CNN is shown in Figure 1.

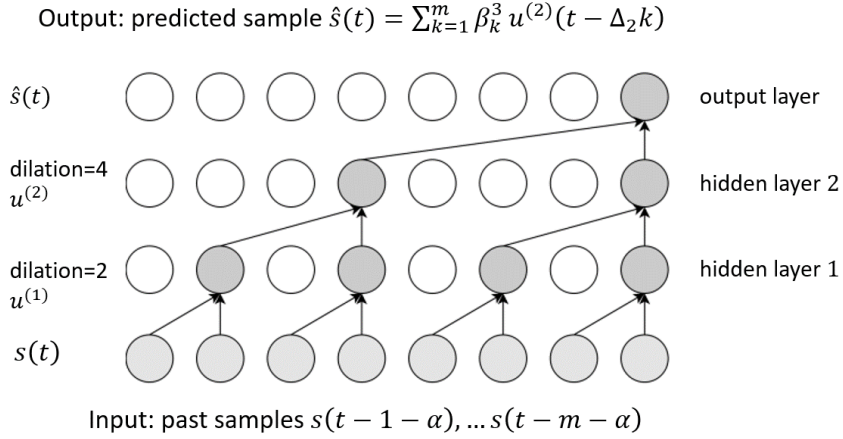

Figure 1. Example of dilated CNN with 3 layers

The predicted values of the signal, derived from employing $L$ layers of the proposed linear dilated CNN, can be expressed as follows:

$$\hat{d}(t) = \hat{s}(t) = \beta^{L-1} *_\Delta \beta^{L-2} *_\Delta \ldots \beta^1 *_\Delta s(t-\alpha), t = 1, \ldots N. \tag{10}$$

Here, each sample $\hat{s}(t)$ is estimated by using $m$ past samples of $s(t)$ with a time lag $\alpha$.

The kernel weights at layer $l$, denoted by $\beta^l$, have a length defined by the kernel size, and the dilation factor at each layer increases by a factor of 2: $\Delta \in [2^0, 2^1, .. 2^{L-1}]$.

The model training proceeds as follows:

1. The signal $s(t)$ is divided into a training set and a validation set, with an 80:20 split ratio.
2. Every sample $s(t)$ is predicted utilizing $m$ past samples of $s(t)$ and a time lag $\alpha$.
3. The mean square error (MSE) between the predicted $\hat{s}(t)$ and actual $s(t)$ values is calculated after each epoch for both the training and validation sets.
4. Early stopping is used to prevent overfitting, stopping the training process when the running mean of the validation set MSE no longer improves.

Further details on the method and its benefits can be found in [21]. The method offers a solution to the PSS problem, resulting in the separation of signal $s(t)$ into its constituent components $d(t)$ and $x(t)$. The following section discusses the design of the deconvolution filter solely employing the stochastic component $x(t)$.

*4.4. Whitening-Based Deconvolution (WBD)*

The stochastic component $x(t)$ includes both the CS2 signal and white noise $c(t) + w(t)$, each subjected to convolution with the TF filter. If $n$ denotes the length of the filter, the elements of the $n \times n$ correlation matrix, $\boldsymbol{R}_{xx}(t,\tau)$, exhibit temporal variation as follows:

$$\boldsymbol{R}_{xx}(t,\tau) = \begin{bmatrix} \sigma_x^2(t) & \cdots & r_x(t, 1-n) \\ \vdots & \ddots & \vdots \\ r_x(t, n-1) & \cdots & \sigma_x^2(t) \end{bmatrix}, \tag{11}$$

where $r_x(t,\tau) = \mathbb{E}\{x(t)x(t-\tau)\}$ and $\sigma_x^2(t)$ is the time-varying variance of $x(t)$. Due to the TF effect, $\boldsymbol{R}_{xx}(t,\tau)$ is not diagonal, and the off-diagonal elements are time-dependent, determined solely by the variance of the CS2 signal $\sigma_c^2(t)$, as demonstrated in [34]. The matrix $\boldsymbol{R}_{xx}(t,\tau)$ can be represented as the product of the time-dependent function $\sigma_c^2(t) + \sigma_w^2$ and a time-independent matrix depending on the time lag $\tau$, as follows:

$$\boldsymbol{R}_{xx}(t,\tau) = (\sigma_c^2(t) + \sigma_w^2)\begin{bmatrix} \varphi(0) & \cdots & \varphi(1-n) \\ \vdots & \ddots & \vdots \\ \varphi(n-1) & \cdots & \varphi(0) \end{bmatrix} \equiv (\sigma_c^2(t) + \sigma_w^2)\boldsymbol{\Phi}(\tau). \tag{12}$$

Here, the elements $\varphi(\tau)$ represent multiplications among various TF coefficients, depending solely on $\tau$, as follows:

$$\varphi(\tau) \equiv \sum_{i-j=\tau} h_i h_j, \quad \forall i,j \in [1,\ldots,n]. \tag{13}$$

The deconvolution filter should remain constant to correspond to the inverse of the constant TF. When the elements of $\boldsymbol{R}_{xx}(t,\tau)$ are estimated using a time frame of length $N$, which is significantly larger than the period $T$, denoted as $N \gg T$, the time-dependent function $\sigma_c^2(t)$ can be expressed as the sum of the average $\bar{\sigma}_c^2$ and the time-varying component $\tilde{\sigma}_c^2(t)$ as follows:

$$\boldsymbol{R}_{xx}(t,\tau) \underset{N \gg T}{=} \underbrace{(\bar{\sigma}_c^2 + \sigma_w^2)\boldsymbol{\Phi}(\tau)}_{\overline{\boldsymbol{R}}_{xx}(\tau)} + \underbrace{\tilde{\sigma}_c^2(t)\boldsymbol{\Phi}(\tau)}_{\widetilde{\boldsymbol{R}}_{xx}(t,\tau)} \equiv \overline{\boldsymbol{R}}_{xx}(\tau) + \widetilde{\boldsymbol{R}}_{xx}(t,\tau)$$

$$\bar{\sigma}_c^2 = \frac{1}{N}\sum_{t=1}^{N} \sigma_c^2(t),$$

(14)

where $\bar{\sigma}_c^2$ is the average of $\sigma_c^2(t)$ over the time frame $N \gg T$.

The deconvolution filter constructed using $\overline{\boldsymbol{R}}_{xx}(\tau)$ as in the first part of equation (14) will be equivalent to the filter required for $\widetilde{\boldsymbol{R}}_{xx}(t,\tau)$. This is true because both parts depend on the matrix $\boldsymbol{\Phi}(\tau)$, which is solely determined by the coefficients of the constant TF the deconvolution filter aims to eliminate. Hence, a constant deconvolution filter $g$ exists and can be designed to render $x(t)$ zero-mean and uncorrelated as required in definition 2.2. As proposed in [34], the zero-phase filter can be estimated using Mahalanobis whitening [33] applied to $\overline{\boldsymbol{R}}_{xx}(\tau)$ by converting it into a diagonal matrix and keeping the signal energy unchanged:

$$\boldsymbol{R}_{yy}(\tau) = \boldsymbol{G}\boldsymbol{R}_{xx}(\tau)\boldsymbol{G}^{\mathrm{T}} = \bar{\sigma}_x^2 \boldsymbol{I}$$

$$\boldsymbol{G} = \bar{\sigma}_x \overline{\boldsymbol{R}}_{xx}^{-1/2}(\tau), \tag{15}$$

$$\hat{g} = \boldsymbol{G}\boldsymbol{i}_{(n-1)/2}$$

where $G$ and $I$ represent the $n \times n$ whitening transformation and the unit diagonal matrices, respectively, while $i_{(n-1)/2}$ denotes the $n \times 1$ basis vector with a value of 1 at the $(n-1)/2$th row and zero otherwise.

*4.5. Deconvolution*

By using $\hat{d}(t)$ and $\hat{x}(t)$, estimated in equation (8), the original periodic component $p(t)$, phase-locked to shaft and gear vibrations, and the signal $y(t)$, encompassing bearing fault and noise, are estimated by applying the estimated deconvolution filter $\hat{g}$ as follows:

$$\hat{p}(t) = \hat{d}(t) * \hat{g}$$
$$y(t) \equiv \hat{x}(t) * \hat{g} \quad (16)$$

Consequently, the true waveform of the periodic component $p(t)$ is estimated by employing the deconvolution filter $\hat{g}$, built by utilizing the stochastic signal component, assuming both components underwent the same TF. The estimation of the waveform $q(t)$ using the signal $y(t)$ is presented in the subsequent section.

*4.6. Distributed Bearing Fault Waveform*

Given that the variance $\sigma_w^2$ is known and assuming $\sigma_e^2 = 1$ without loss of generality, the non-negative function $q(t)$ can be estimated as follows:

$$\mathbb{E}\{y^2(t)\} = q^2(t)\sigma_e^2 + \sigma_w^2,$$
$$\hat{q}(t) = \sqrt{\mathbb{E}\{y^2(t)\} - \sigma_w^2}. \quad (17)$$

However, in practice, $\sigma_w^2$ is unknown and estimating of $\mathbb{E}\{y^2(t)\}$ becomes challenging due to the lack of information about its temporal variations.

As suggested in [34], $\mathbb{E}\{y^2(t)\}$ can be estimated in the frequency domain by assuming the periodic properties of $q(t)$. However, before this estimation, a hypothesis test should be conducted to confirm the presence of the CS2 component in $y(t)$. The solution presented here is inspired by the work in [38], which showed that the squared Discrete Fourier Transform (DFT) coefficients of the log envelope, $log(y^2(t))$, follow a chi-square distribution with a non-central parameter that does not depend on the variance of the noise, $\sigma_w^2$. Specifically, the squared amplitude exhibits a central chi-square distribution in frequencies related to noise and a non-central distribution in frequencies associated with the CS2 component:

$$|Y_L(f)|^2 \equiv \left| F\left( log(y^2(t)) \right) \right|^2, f \in [0, \dots, f_s]$$
$$\frac{|Y_L(f)|^2}{\pi^2/4N} \sim \chi_2^2(\lambda) \quad (18)$$
$$\lambda = \begin{cases} = 0, & f \in noise \\ > 0, & f \in CS2 \end{cases}$$

where $F( )$ denotes the DFT operator, $N$ represents the DFT length, and $\chi_2^2(\lambda)$ signifies the chi-square distribution with 2 degrees of freedom and the non-central parameter $\lambda$.

The presence of CS2 can be detected by conducting a hypothesis test for each frequency:
$H_0$: "The signal $y(t)$ does not contain a CS2 component at frequency $f$"
$H_1$: "The signal $y(t)$ contains a CS2 component at frequency $f$"
If $|Y_L(f)|^2/4N\pi^2 > \chi_2^2(1-p)$, then $H_1$ is accepted; otherwise, $H_0$ is accepted at frequency $f$. Here, the parameter $p$ represents the desired percentile.

The outcome of the hypothesis test significantly relies on selecting the percentile that defines the threshold. To streamline this process working autonomously on sensor, the $|Y_L(f)|^2$ can be estimated after segmenting $y(t)$ into frames and averaging the squared DFT coefficients of these frames, prior to conducting the hypothesis test. In this scenario, the distribution of $|Y_L(f)|^2$ conforms to a Gamma distribution with parameters determined by the number of averages:

$$|Y_L(f)|^2 = \frac{1}{K}\sum_{k=1}^{K}\left|Y_L^{(k)}(f)\right|^2 \quad (19)$$

$$|Y_L(f)|^2 \sim \Gamma(\alpha = K, \beta = 2/K)$$

where $k$ denotes the frame number, $\Gamma(\alpha = K, \beta = 2/K)$ represents a Gamma distribution with parameters $\alpha, \beta$ uniquely determined by the number of averages $K$. The hypothesis test will use thresholds derived from the Gamma distribution. Examples of thresholds for $p = 0.01$ and $p = 0.001$ are provided in Table 1.

Table 1 Hypothesis test thresholds according to the Gamma distribution

| Number of averages | $p = 0.01$ | $p = 0.001$ |
|---|---|---|
| 3 | 5.60396 | 7.48591 |
| 5 | 4.64185 | 5.91766 |
| 10 | 3.75662 | 4.53147 |
| 20 | 3.18454 | 3.32024 |
| 40 | 2.80822 | 3.12098 |

The difference between the percentiles diminishes as the number of averages approaches 20 (shaded in gray). For example, a single threshold of 3.32 can be used as generic threshold, accommodating percentiles between 0.01 and 0.001. However, augmenting the number of averages reduces frequency resolution, as it employs smaller time frames for the same signal length. We recommend employing $K \leq 20$ averages as a trade-off between straightforward thresholding, computational complexity, and frequency resolution.

If the set of frequencies $\{f_i\}$ satisfying the hypothesis $H_1$ is empty, then the CS2 component doesn't exist in $y(t)$, indicating that there is no need for estimating $q(t)$ since the bearing fault, if exists, is deeply embedded in noise. Otherwise, we proceed to estimate $q(t)$ using the DFT of $y^2(t)$. After setting all the DFT coefficients except those corresponding to $\{f_i\}$ to zero, we return to the time domain using the inverse DFT:

$$Y^q(f) \equiv \begin{cases} 0, & f \notin \{f_i\} \\ Y(f), & f \in \{f_i\} \end{cases} \quad (20)$$

$$\hat{q}(t) \cong \sqrt{\mathbb{E}\{y^2(t)\}} = \sqrt{F^{-1}(Y^q(f))}$$

where $Y^q(f)$ denotes the modified DFT of $y^2(t)$ and $F^{-1}()$ represents the inverse DFT. Since the variance of noise $\sigma_w^2$ is unknown, the estimation of $q(t)$ is biased up to a DC factor.

### 4.7. Local Bearing Fault Waveform

The local fault waveform (as defined in equation (2)) can be estimated in a similar way assuming the variance between different shock appearances $\Delta T$ is small when the operating conditions are stable. In this case, the local fault signal is equivalent to the distributed one re-sampled with sampling frequency of $f_s \cong 1/\Delta T$ as follows:

$$y(t) \cong q(k\Delta T)e(k\Delta T) + w(t), k \in \mathbb{N} \quad (21)$$

$$\mathbb{E}\{y^2(t)\} = q^2(k\Delta T)\sigma_e^2 + \sigma_w^2.$$

If the variance of $\Delta T$ is small, then the $y^2(t)$ in the local fault case represents an aliased version of the distributed fault and noise. Since the local fault spectrum includes the DFT peaks of the $q^2(k\Delta T)$, then its estimation may be performed the same way as that of the distributed fault assuming $q(t)$ is composed of the finite number of cyclic elements. Moreover, in practice both fault types can be roughly monitored by trending the time history of $F(y^2(t))$ amplitudes at frequencies $\{f_i\}$ answering the $H_1$ hypothesis.

*4.8. Algorithm Summary*

The table below summarizes the method limiting it by the distributed bearing faults.

Table 2 Algorithm Summary

| # | Algorithm Steps |
|---|---|
|   | Receive $N$ measurements $\{s(1), \dots s(N)\}$ to be used in the estimation of $p(t)$ and $q(t)$. |
| 1 | Estimate the deterministic component $\hat{s}(t)$ using dilated CNN (Section 4.3) |
| 2 | Estimate the stochastic component: $x(t) = s(t) - \hat{s}(t)$ |
| 3 | Build matrix $\boldsymbol{R}_{xx}$:<br>$$\boldsymbol{R}_{xx} = \begin{bmatrix} r_x(0) & \cdots & r_x(1-n) \\ \vdots & \ddots & \vdots \\ r_x(n-1) & \cdots & r_x(0) \end{bmatrix}$$ $$r_x(\tau) = \frac{1}{N}\sum_{t=1}^{N} x(t)x(t-\tau), \tau = 0,1,\dots,n$$ |
| 4 | Estimate the deconvolution filter as the middle column of matrix $\boldsymbol{G}$:<br>$$\boldsymbol{G} = \left(\sum_{t=1}^{N} x(t)^2\, \boldsymbol{R}_{xx}^{-1}\right)^{1/2}$$ $$\hat{g} = \boldsymbol{G} i_{(n-1)/2}$$ |
| 6 | Perform the deconvolution and estimate $p(t)$:<br>$$\hat{p}(t) = \hat{d}(t) * \hat{g}$$ |
| 7 | Perform the deconvolution of $x(t)$:<br>$$y(t) = \hat{x}(t) * \hat{g}$$ |
| 8 | Define $K$ and calculate the averaged spectrum of log envelope (default $K = 20$):<br>$$|Y_L(f)|^2 = \frac{1}{K}\sum_{k=1}^{K} \left|Y_L^{(k)}(f)\right|^2$$ |
| 9 | Perform a hypothesis test for the existence of CS2 at each frequency, following the definition of the percentile $p$ (default $p = 0.001$)<br>$$\begin{cases} f_i \in H_0, & |Y_L(f_i)|^2 \leq \Gamma\left(K,\frac{2}{K}\right)(1-p) \\ f_i \in H_1, & |Y_L(f_i)|^2 > \Gamma\left(K,\frac{2}{K}\right)(1-p) \end{cases}$$ |

| 10 | If there are no frequencies where $H_1$ is true, then the CS2 component does not exist. Exit. |
| --- | --- |
| 11 | Otherwise, detect the set of frequencies $\{f_i\}$ where $|Y_L(f)|^2$>*thresh* (default *thresh*=3.32 for $K = 20$) |
| 12 | Modify $Y_L(f)$ given the set of frequencies $\{f_i\}$: $$Y_L^q(f) = \begin{cases} 0, & f \notin \{f_i\} \\ Y_L(f), & f \in \{f_i\} \end{cases}$$ |
| 13 | Estimate $q(t)$: $$\hat{q}(t) = \sqrt{F^{-1}(Y^q(f))}$$ |

## 5. Experimental Results

### 5.1. Simulations

The simulations were designed to assess the method's robustness in relation to four parameters: the number of poles in the TF, the number of cyclic components in $p(t)$ and $q(t)$, and the method's robustness to noise $w(t)$. To prevent the accuracy of $p(t)$'s estimation from influencing $q(t)$'s estimation, these two processes were performed independently, with separate noise generation. The simulation involved 200 iterations of the algorithm for each parameter combination, with 100 iterations dedicated to $p(t)$ and another 100 for $q(t)$. The need for separate runs for $p(t)$ and $q(t)$ arose from their differing Signal-to-Noise Ratio (SNR) definitions, as will be further explained in this section.

One second signals $s(t)$ consisting of $N = 24{,}000$ samples in each run were generated using equation (6):

$$s(t) = \big(p(t) + c(t) + w(t)\big) * h, \qquad t = \frac{1}{N}[0, \dots, N-1]$$

$$p(t) = \sum_{v=1}^{V} A_v \cos(\omega_v t + \varphi_v). \tag{22}$$

The local and distributed bearing faults are simulated randomly during the runs to check the robustness of the algorithm to the different fault types. The simulation was performed as follows:

$$q(t) = \sum_{l=1}^{L} \big(1 + B_l \cos(\omega_l t + \varphi_l)\big),$$

$$Local: c(t) = q(t) \sum_{k} e(t)\delta(t - T_k),$$

$$Distributed: c(t) = q(t)e(t),$$

$$e(t) \sim \mathcal{N}(0,1), \qquad w(t) \sim \mathcal{N}(0, \sigma_w^2), \qquad T_{k+1} - T_k \sim \mathcal{N}(0, \sigma_T^2)$$

The simulation settings were adjusted prior to each run as follows:
- The number of poles in the TF, which define the coefficients $h_i$ and their length $n$, varied between 5 and 20. The frequency of poles was uniformly distributed on the whole frequency range.
- The number of cyclic components $V$ in the gear-related component $p(t)$, used to simulate dominant gear vibrations, ranged from 5 to 50. The frequencies $\omega_v$ and phases $\varphi_v$ were uniformly distributed within the ranges $(0, \pi)$ and $(-\pi, \pi)$ respectively.
- The number of cyclic components $L$ in the bearing fault, used to simulate local and distributed bearing faults, varied between 5 and 10. The frequencies $\omega_l$ and phases $\varphi_l$ were uniformly distributed within the ranges $(0, \pi)$ and $(-\pi, \pi)$ respectively.
- The variance of noise $e(t)$ was set to 1 with a mean of zero.
- The variance of noise $w(t)$ was set to 1, and the energy of $p(t)$ and $q(t)$ was adjusted to achieve the required SNR.
- The variance $\sigma_T^2$ was equal to 0.01 sec.
- The SNR, defined as $SNR = 10\log 10 \frac{\|p(t)\|_2^2}{\|q(t)\|_2^2 + \sigma_w^2}$ for $p(t)$ and $SNR = 10\log 10 \frac{\|q(t)\|_2^2}{\sigma_w^2}$ for $q(t)$, was specified separately for each run depending on whether the estimation was for $p(t)$ or $q(t)$. The SNR varied between -20dB and 20dB.

The new method's performance was evaluated using the Coefficient of Determination $R^2$ for estimation of both $p(t)$ and $q(t)$:

$$R^2 = 1 - \frac{\sum_{t=1}^{N}(x(t)-\hat{x}(t))^2}{\sum_{t=1}^{N}(x(t)-\bar{x})^2}, \quad (23)$$

where $N = 24{,}000$ represents the length of the signal, $x(t)$ denotes the true value, $\hat{x}(t)$ is the estimated value, and $\bar{x}$ is the average of the true signal.

An example of the generated signal $s(t)$ is demonstrated in Figure 2. The component $p(t)$ comprises $K = 5$ components with an SNR of 0dB. The component $q(t)$ consists of $L = 5$ cyclic components.

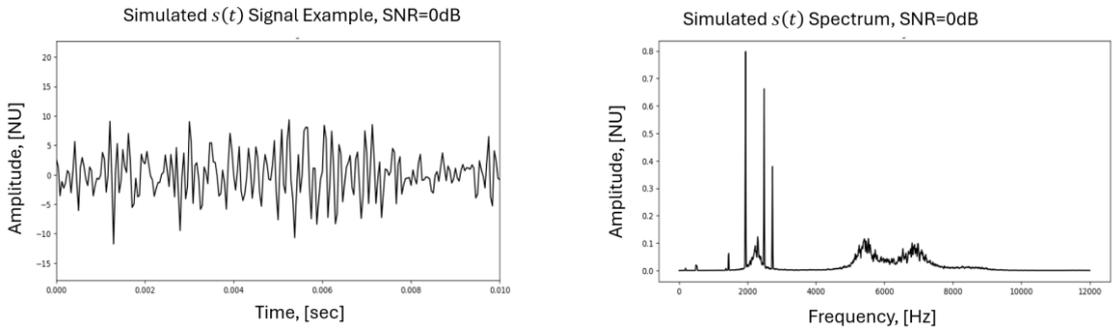

Figure 2. An example of a simulated signal $s(t)$ (left) and its corresponding spectrum (right) with $K = 5$ components within $p(t)$ and $L = 5$ components within $q(t)$.

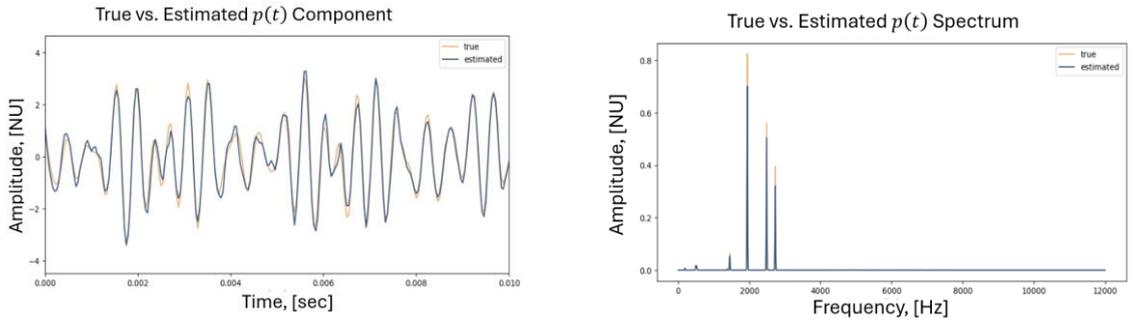

Figure 3. An example of estimated gear-related component $p(t)$ (left) and its spectrum (right) with $K = 5$ cyclic components

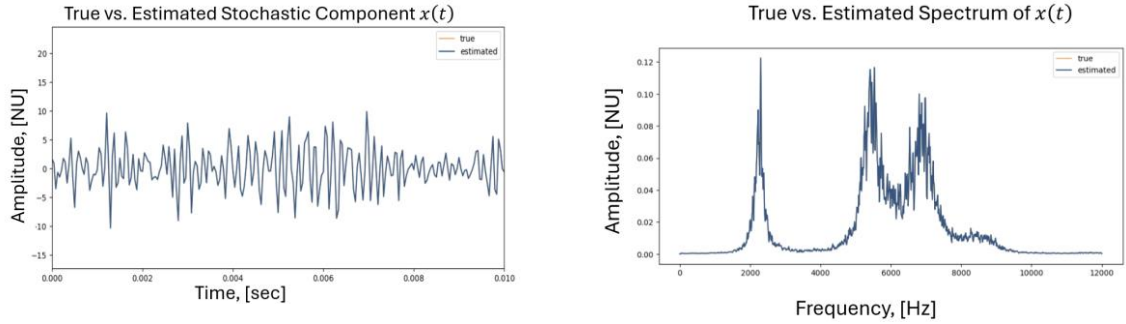

Figure 4. An example of estimated stochastic component $x(t)$ (left) and its spectrum (right) with $K = 5$ sinusoids of $p(t)$ and $L = 5$ sinusoids of $q(t)$.

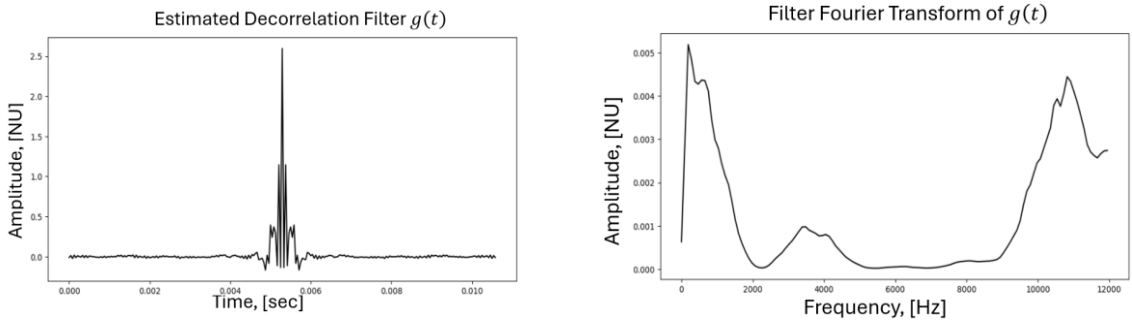

Figure 5. An example of deconvolution filter (left) and its spectrum (right) where the number of poles in TF was equal to 5.

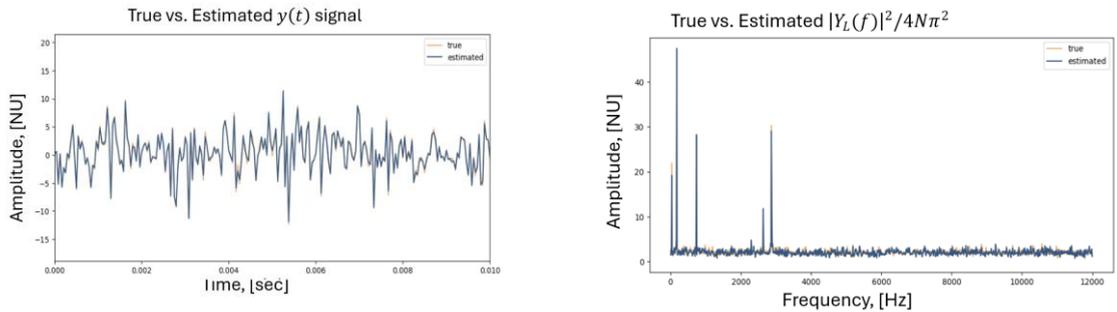

Figure 6. An example of component $y(t)$ (left) and the normalized spectrum of its log envelope spectrum (right). The true signal is blue and the estimated is orange.

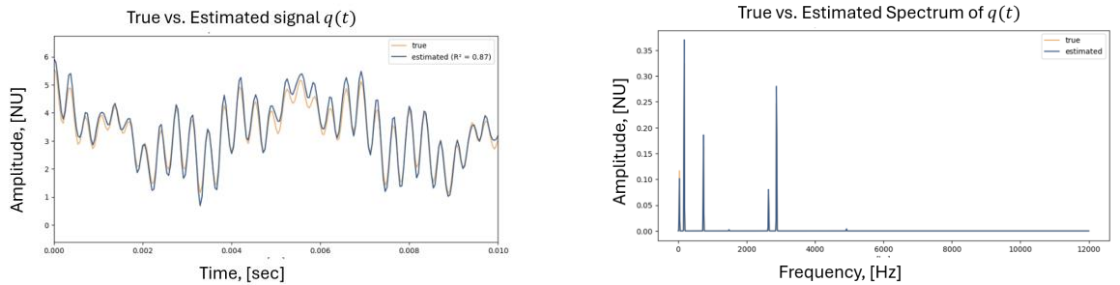

Figure 7. An example of estimated component $q(t)$ (left) and its spectrum (right). The true signal is blue and the estimated is orange.

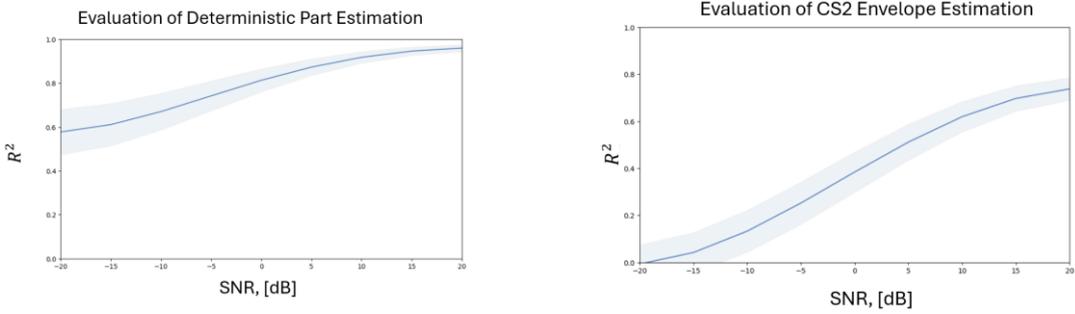

Figure 8. Coefficient of Determination $R^2$ vs. SNR evaluating estimation of $p(t)$ and $q(t)$. The spread around the graphs corresponds to ±1 standard deviation.

*5.2. Error Analysis, Parameters Choice, and Method Limitations*

The simulations validate the method's resilience across a spectrum of Signal-to-Noise Ratio (SNR) scenarios. The estimation of $p(t)$ exhibits notable robustness to noise. Conversely, the estimation of $q(t)$ demonstrates comparatively less resilience, as the identification of its cyclic components' frequency peaks necessitates surpassing a threshold, which proves challenging under low SNR conditions. These results reflect different combinations of TF poles and cyclic components in $q(t)$ and $p(t)$ respectively, suggesting that the method's performance remains unaffected by signal parameters.

There are several parameters that need to be determined without prior knowledge of the monitored system:
  a. The length $n$ of the TF and the deconvolution filter $g$ are among the parameters to be determined. The length $n$ represents a balance between estimation accuracy, as defined by the deconvolution filter estimation, and computational cost. A higher $n$ promises the necessary filter length and its frequency resolution, but increases computational complexity by expanding the size of $\boldsymbol{R}_{xx}$. Consequently, inverting this enlarged $\boldsymbol{R}_{xx}$ becomes more computationally intensive."
  b. The parameters of the dilated CNN include the reception field $m$, training size, kernel size, and model depth. Since model training occurs on a single recording, the training size is primarily determined by the signal length $N$. The reception field is determined by the inverse of the desired frequency resolution; for example, a reception field of 0.5 seconds corresponds to a frequency resolution of 2 Hz, which is suitable for reliable condition monitoring of most industrial equipment.
  c. The kernel size is uniquely determined by $m$ and the model depth, as demonstrated in [21]. Increasing the model depth reduces the number of optimization parameters and shortens the required signal length for training. However, deeper models consume more computational resources during optimization, which may be limited in battery-powered wireless sensors. We believe that the optimal trade-off for industrial vibrations processing lies within four levels.
  d. The large time-lag $\alpha$ promises that the correlation of the stochastic component is zero for time lags $|\tau| > \alpha$. Given that the correlation is defined mainly by the length of the TF impulse response, once $n$ is determined, selecting $\alpha \geq n$ ensures zero correlation of the stochastic component, as per our definition.

There are several limitations which should be considered when using the method:
  a. In our study the vibrations signals are assumed to be generated under constant rotating speed and loads which is not true in practice. In practice the information about the operating conditions is unavailable on sensor. The signal validation can be performed by checking stability of signal energy within the recording. Since gear- or shaft- related harmonics generate the major energy, its stability should promise rotational speed stability too. The load stability may be guaranteed by working with small recordings assuming that

  the system inertia varies slowly. Further research is required to extend the method for varying operating conditions.
b. The method is limited to a specific type of CS2 signals which should model bearing faults. In practice the signals can be slightly different and increase the estimation errors.
c. The TF is assumed to be invertible. If not, the deconvolution filter in Section **Error! Reference source not found.** may be designed by using the extended subspace whitening like suggested in [39] which is not in scope of the current study.
d. In practical scenarios, the measurement setup may also incorporate additional white noise added after the signals traverse the TF. Further the noise may be amplified by the deconvolution filter, requiring further work to rectify this issue.

*5.3. Bearing Fault Detection in Wind Turbines*

 To assess a new method using real-world data from actual machinery, we utilized vibrations from an offshore 5MW wind turbine. Each vibration recording encompassed signals from both the low-speed gear and the bearings. Our aim is to showcase the ability of the new method to detect bearing faults at early stages when executed directly on the sensor, without any prior knowledge of the system or external measurements. The signals were collected using a measurement system developed by RSL Electronics and deployed across numerous locations worldwide.

 One-second vibration signals were recorded using industrial accelerometers with 24KHz sampling frequency from the low-speed stage of the gearbox, as illustrated in Figure 9.

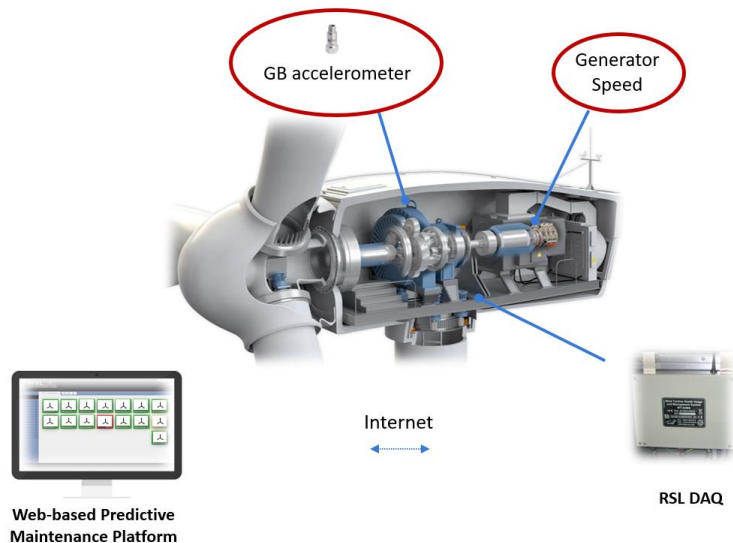

Figure 9. Wind turbine measurement system and sensor locations

 We assume that no information regarding resonances or bearing fault frequencies is available at the sensor. Therefore, classical methods such as bandpass filtering and envelope detection are not applicable in this context. Instead, the CS2 waveform detection, as proposed in our study, is employed. To ensure data quality, only recordings meeting the energy stability criterion were included. This criterion was defined as a maximum change of 5% in signal root mean square (rms) within a recording divided into 20 frames. As outlined in the methodology section, two components $p(t)$ and $q(t)$, were estimated for each recording. The algorithm configuration is detailed in Table 3.

Table 3 Algorithm configuration

| Parameter | Name | Value |
|---|---|---|
| Recording length | $N$ | 24000 |
| TF filter length | $n$ | 250 |
| Time lag | $\alpha$ | 500 |
| Number of averages | $K$ | 20 |
| CS2 hypothesis threshold | $thresh$ | 3.2 |
| Dilated CNN: | | |
|     Receptive field | $m$ | 2011 |
|     Depth | | 4 |
|     Kernel size | | 135 |
|     Train/test partition | | 80/20 |

An example of applying the method to the vibrations recorded from the low-speed gear of the wind turbine is shown below.

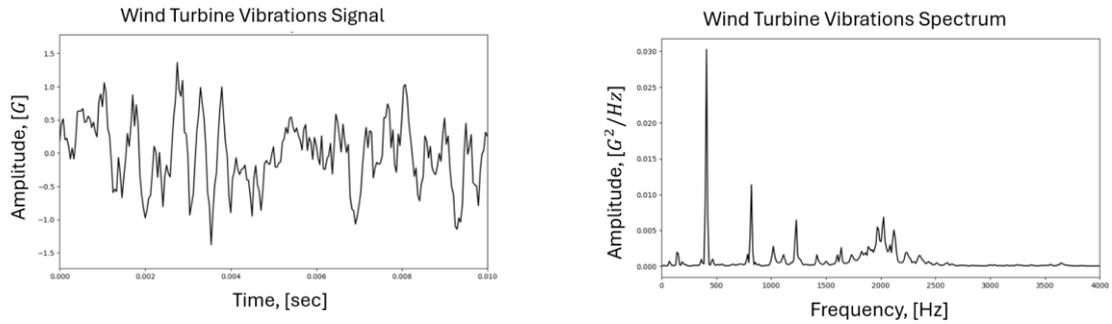

Figure 10. An example of the signal $s(t)$ (left) and its corresponding spectrum (right) measured from low-speed gear of wind turbines.

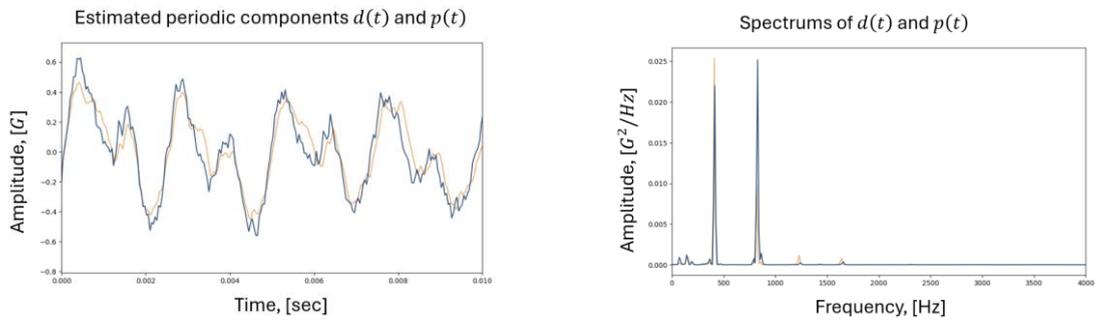

Figure 11. An example of estimated periodic part $d(t)$ (left orange), gear-related component $p(t)$ (left blue) and their corresponding spectrums (right).

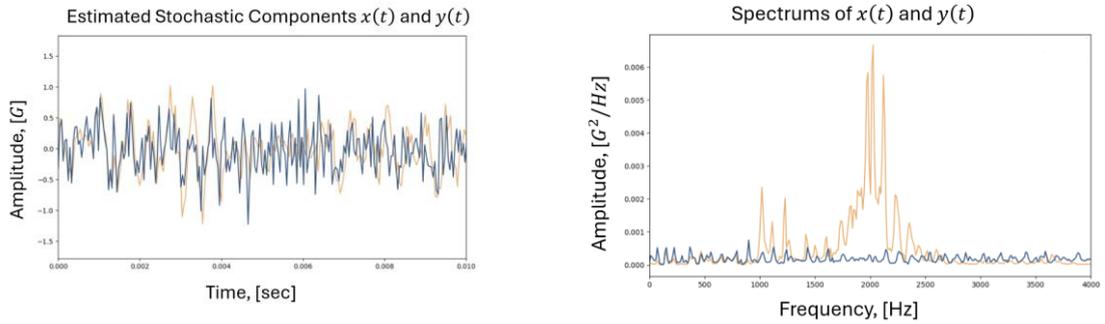

Figure 12. An example of estimated components $x(t)$ and $y(t)$ (left orange and left blue respectively) and their corresponding spectrums (right).

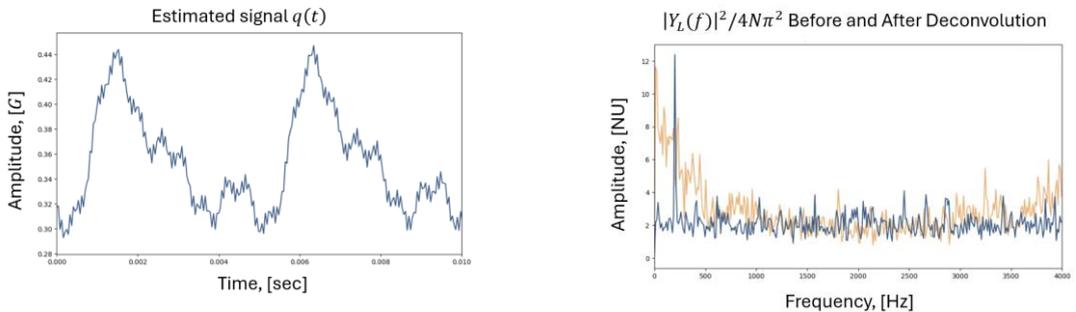

Figure 13. An example of estimated $q(t)$ (left) and its normalized log envelope spectrum (right blue). The log envelope spectrum before the deconvolution is orange.

## 6. Discussion

Figure 10 to Figure 13 present an example of using the method when analyzing real vibrations from wind turbine where no additional information exists. This example demonstrates the capability to recognize bearing faults at early stages. The peak at Figure 13 corresponds to the bearing outer race fault and is clearly emphasized when using the method vs. the standard envelope analysis. The method automatically separates the gear and bearing sources without relying on prior knowledge of the system or external measurements. Notably, it highlights early-stage bearing faults in the absence of external information. However, it does not differentiate between local and distributed bearing faults since both may have peaks in the spectrum. Additionally, the method necessitates stable operational conditions, which are challenging to ascertain without external measurements. We addressed this by using signal energy stability as a rough estimate of stability. Moving forward, the proposed method should be extended to accommodate varying conditions as well.

## 7. Conclusions

In this study, we introduced an innovative method for estimating time waveforms of two primary vibration sources: gear-related and bearing-related, without relying on external measurements or prior knowledge of the system. This capability enables the method to be directly implemented on sensors. Our approach demonstrates robustness to noise and to variations in signal and monitored system characteristics. The waveform estimation is tailored to signals associated with distributed and localized bearing faults. Further research is warranted to expand the method's applicability to practical scenarios, such as variable operating conditions and the presence of additional noise sources. Additionally, exploring the potential of this method to enhance machine learning training is promising, particularly as a preprocessing step when aggregating training data from identical units with differing time-frequency characteristics.

**Declaration of competing interest**

The authors declare that they have no known competing financial interests or personal relationships that could have appeared to influence the work reported in this paper.